\pdfoutput=1

\documentclass[11pt,a4paper]{article}
\usepackage[T1,LGR]{fontenc} 
\usepackage[greek,english]{babel}

\usepackage{textgreek}
\usepackage{enumitem}
\usepackage{tabularx}
\usepackage[margin=1in, bottom=1in]{geometry}
\usepackage{soul}
\usepackage{quoting}

\usepackage{multirow}
\usepackage{array, booktabs, colortbl, xcolor, caption}
\usepackage{times, latexsym, amsmath}

\usepackage{booktabs}
\usepackage{multirow}
\usepackage{colortbl}
\usepackage{array}

\usepackage{latexsym}
\usepackage{graphicx}
\usepackage{subfigure}

\usepackage{algorithm}
\usepackage{algpseudocode}
\usepackage{amsmath, amssymb}

\usepackage[preprint]{acl}

\usepackage{times}
\usepackage{latexsym}
\usepackage{amsmath}
\usepackage{amssymb}

\usepackage[T1]{fontenc}
\usepackage[utf8]{inputenc}

\usepackage{microtype}

\usepackage{inconsolata}

\usepackage{graphicx}
\usepackage{csquotes}

\usepackage{hyperref}
\addto\extrasenglish{
    
}

\newcommand{\puEmoji}{\includegraphics[width=0.8em]{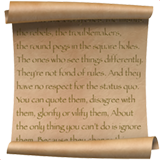}}
\newcommand{\mitEmoji}{\includegraphics[width=0.8em]{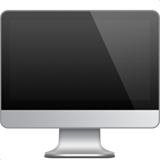}}
\newcommand{\huEmoji}{\includegraphics[width=0.8em]{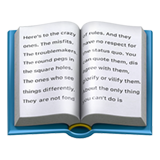}}

\title{An Annotated Dataset of Errors in Premodern Greek \\and Baselines for Detecting Them}

\author{Creston Brooks\thanks{CB and JH contributed equally as first authors.}\thanks{Correspondence addressed to.}\puEmoji\quad Johannes Haubold\footnotemark[1]\puEmoji\quad \\ \textbf{Charlie Cowen-Breen\footnotemark[2]\mitEmoji}\quad \textbf{Jay White\puEmoji}\quad
\textbf{Desmond DeVaul\puEmoji}\quad \\ \textbf{Frederick Riemenschneider\huEmoji}\quad \textbf{Karthik Narasimhan\puEmoji}\quad \textbf{Barbara Graziosi\puEmoji} \vspace{0.1cm} \\
  \puEmoji \hspace{0.1cm}  Princeton University, \hspace{0.1cm}  \huEmoji \hspace{0.1cm} 
 Heidelberg University, \hspace{0.1cm}  \mitEmoji \hspace{0.1cm}  MIT      \\
  \texttt{cabrooks@princeton.edu}, \texttt{ccbreen@mit.edu}
}

\begin{document}
\maketitle

\begin{abstract}
 As premodern texts are passed down over centuries, errors inevitably accrue. These errors can be challenging to identify, as some have survived undetected for so long precisely because they are so elusive. While prior work has evaluated error detection methods on \textit{artificially-generated} errors, we introduce the first dataset of \textit{real} errors in premodern Greek, enabling the evaluation of error detection methods on errors that genuinely accumulated at some stage in the centuries-long copying process. To create this dataset, we use metrics derived from BERT conditionals to sample 1,000 words more likely to contain errors, which are then annotated and labeled by a domain expert as errors or not. We then propose and evaluate new error detection methods and find that our discriminator-based detector outperforms all other methods, improving the true positive rate for classifying real errors by 5\%. We additionally observe that scribal errors are more difficult to detect than print or digitization errors. Our dataset enables the evaluation of error detection methods on real errors in premodern texts for the first time, providing a benchmark for developing more effective error detection algorithms to assist scholars in restoring premodern works.
\end{abstract}


\section{Introduction}
Ancient texts have been passed down over hundreds of years.
The oldest surviving manuscripts of Sophocles, Plato, and Aristotle date to the ninth and tenth centuries CE, long after the original works were composed in the fifth and fourth centuries BCE. Thus, what is left to us today are copies of copies of copies. Throughout this process of copying, errors have accumulated in three main ways:


\begin{description}[leftmargin=0cm]
\setlength\itemsep{0.2em}
    \item[Scribal errors:] Scribes copying manuscripts over centuries introduce changes—such as adding, omitting, repeating, or simplifying text—that go unnoticed by subsequent scribes and are then copied forward as though they were the original text.
    
    \item[Print errors:] Modern scholars occasionally misread manuscripts or introduce typos when creating  editions, leading to mistakes in published versions.
    \item[Digitization errors:] The conversion of printed texts to online versions, whether through manual typing or automated processes, introduces additional errors.
\end{description}

\begin{figure}
    \centering
    \includegraphics[width=1.0\linewidth]{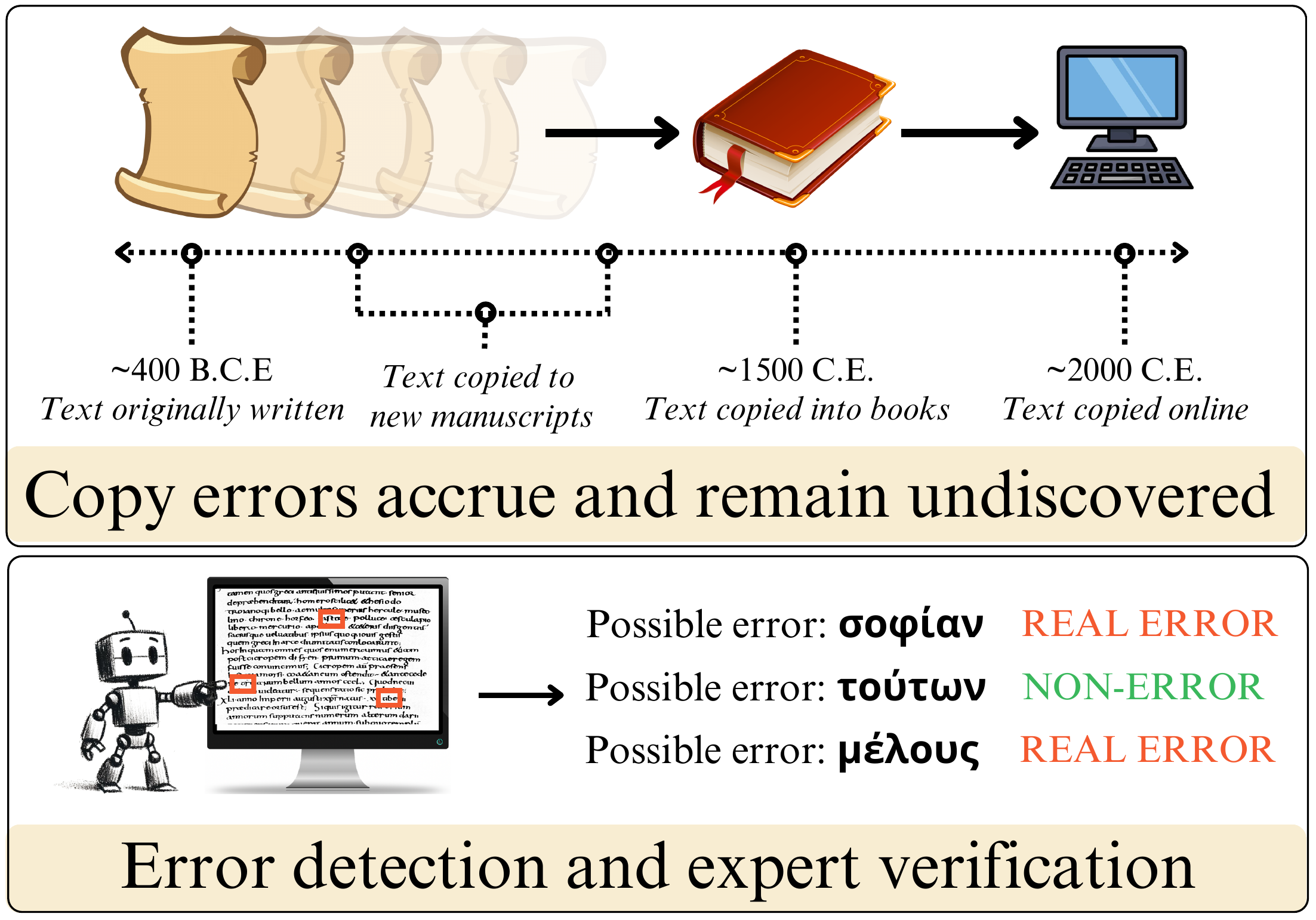}
    \caption{Errors in premodern texts accumulate over centuries of copying. Using machine-learning methods and expert labeling, we create the first dataset of real errors in premodern Greek texts.}
    \label{fig:front_page}
\end{figure}

Errors made at all stages, from the earliest copies of an ancient text to what we read online today, threaten the faithful preservation of that text, change its original wording, and impede our understanding of it. The most insidious errors are not simple typos, but alterations that make logical sense, allowing them to persist undetected.

Only one unsupervised method has been proposed for detecting errors in premodern texts using machine-learning techniques: \citet{cowen-breen-etal-2023-logion} directly leverage distributions learned by a BERT model \citep{devlin-etal-2019-bert} without task-specific
fine-tuning. This method, while successful in identifying a limited number of errors \citep{graziosi2023machine}, has only been broadly evaluated on detecting artificial errors generated by random character replacement. 

Until now, there has been no available dataset of errors that resulted from the natural process of copying illustrated in \autoref{fig:front_page}. In this paper, we introduce the first expert-labeled dataset of real errors (scribal, print, and digitization), enabling the evaluation of error detection methods on real errors rather than artificial ones. We use a form of automated over-sampling to select potential errors, which a domain expert then spends over 100 hours labeling (see \autoref{sec:dataset}).

Using this dataset, we evaluate \citeposs{cowen-breen-etal-2023-logion} existing error detection method and propose new unsupervised methods, including one inspired by protein engineering and another using an ELECTRA discriminator \citep{clark2020electra}. We also establish a large language model (LLM) baseline with few-shot prompting using GPT-3.5 and GPT-4 \citep{achiam2023gpt}. The ELECTRA discriminator improves the true positive rate over the next best method by 5\%, while GPT-3.5 and GPT-4 perform only marginally better than random chance, with AUROCs of $0.51$ and $0.57$, respectively. We additionally observe across methods that scribal errors are more difficult to detect than print or digitization errors.

\section{Related Work}
Recent years have seen significant progress in training language models (LMs) on premodern languages including Greek \citep{singh-etal-2021-pilot, yamshchikov-etal-2022-bert, riemenschneider-frank-2023-exploring}. These works make use of various masked language models (MLMs) for tasks such as dependency parsing, lemmatization, and gap infilling. \citet{assael2022restoring} focus on filling gaps in inscriptions, and \citet{jones2022machine} use support vector machines and decision trees to adjudicate between New Testament manuscript variants. \citet{cullhed2024instruct} explores the fine-tuning of modern foundation models for filling gaps in ancient papyri, and \citet{duan-etal-2024-restoring-ancient} take a multimodal approach towards restoring ancient Chinese texts. Notwithstanding these efforts, the field of machine-learning assisted textual restoration remains nascent. 

Other work has focused on the supervised detection and correction of errors introduced by Optical Character Recognition (OCR) and Handwritten Text Recognition (HTR), as opposed to scribal errors and print errors \citep{chiron2017icdar2017, amrhein2018supervised, schaefer-neudecker-2020-two, nguyen2020neural, pavlopoulos-etal-2023-detecting}. Although errors introduced by OCR and HTR can result in garbled text that is challenging to \textit{correct}, they are generally easy to \textit{detect}, since a simple dictionary check can flag nonsensically distorted words. Additionally, these studies largely rely on extensive datasets of OCR/HTR text with aligned ground truth. Many errors we consider (scribal and print) have survived because they often make logical sense and are thus more difficult to detect.

\section{Contributions}
\begin{figure}
    \centering
\includegraphics[width=0.9\linewidth]{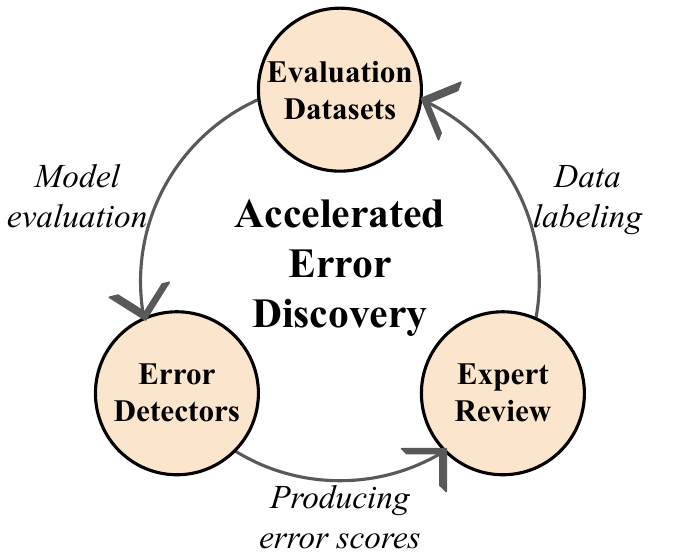}
    \caption{\textbf{Proposed pipeline for accelerated error discovery}. Expert labeling creates evaluation datasets (\autoref{sec:dataset}), leading to better error detectors (\autoref{sec:scoring}), providing higher-quality samples for the next round of expert review.}
    \label{fig:flow_chart}
\end{figure}

Computational textual restoration has previously involved either (i) domain experts using error-detection algorithms to discover a limited number of real errors \citep{graziosi2023machine}, or (ii) broadly evaluating error detection algorithms using datasets of artificially generated errors \citep{SPENCER2004503,Roos:Heikkila:2009,Hoenen:2015a}. In contrast, we introduce the first error detection dataset composed of real errors. We then use this dataset to evaluate the existing error detection method as well as additional methods which we propose. We summarize our contributions as follows:

\begin{enumerate}
    \item We create a dataset of textual errors flagged by machine-learning methods and annotated by a domain expert.\footnote{We make this dataset available, along with the error detectors we evaluate: \url{https://github.com/brooksca3/logion_error_dataset}.}
    \item We propose two new error detection methods: one inspired by protein engineering and another using an ELECTRA discriminator.
    \item We pre-train a suite of models with varying architectures to evaluate the existing and proposed error detection methods using our expert-labeled dataset.
    
  
\end{enumerate}
With real textual problems, labeled and annotated by a domain expert, error detection methods can be effectively evaluated at scale for the first time. In turn, improved error detection capabilities lead to better identification of errors for future domain expert review, propelling the discovery cycle. Here, we enable the cycle of accelerated error discovery seen in \autoref{fig:flow_chart}.

\section{Error Detection}
\label{sec:error_detection_task}

Given a word $w_i$ and its surrounding context $\mathbf{w}=(w_1, \hdots, w_k)$, the task of error detection is to determine whether the given word is an error. More precisely, an \textit{error detector} is a function $T$ such that $T(\mathbf{w},i)$ produces an error score for the word $w_i$ in the given context $\mathbf{w}$.

Error detectors are useful because the scores they produce can yield a list of words deemed most likely to be errors. For example, a word $w_i$ may be shortlisted as a potential error if $T(w,i)>0.99$ for a given detector $T$. Assuming a tolerably successful error detector, words with scores above a certain threshold can be passed on to domain experts for review.

\section{Dataset Creation}\label{sec:dataset}
\subsection{Identifying Real Errors}

We create a dataset of real errors that accumulated as texts were copied first from handwritten manuscripts, then to printed editions, and eventually to digital versions. To do so, we choose the corpus of the 11th-century Byzantine author Michael Psellos, due to its considerable size (1M words) and availability in digitized form. Our domain expert is a philologist who has worked closely with the texts in question \citep{haubold2023konjekturen}.

The rarity of real errors within the corpus means that drawing random words for expert review would be statistically unlikely to yield any positive labels. Additionally, the labeling process is time-consuming, as the domain expert must consult various printed editions, manuscript versions, and, in the case of suspected scribal errors, a range of philological resources.

Therefore, we follow the methodology proposed by \citet{cowen-breen-etal-2023-logion} to over-sample real errors, which we subsequently label:

\begin{itemize}
\item Using a premodern Greek BERT model, we assign a Chance-Confidence Ratio (CCR) score (see \autoref{ccr}) to every word in a subset of the corpus.\footnote{In practice, we randomly divided the text into five parts and presented the top 500 CCR-scoring words from each to the domain expert, who labeled 1,000 words from two parts.}
 
\item We present a list of the 1,000 words with the highest CCR scores to the domain expert who determines whether each word is an error or not. The expert additionally annotates each example with brief philological comments to justify the given label.
\end{itemize}

\begin{figure*}[t]
    \centering
    \includegraphics[width=1.0\linewidth]{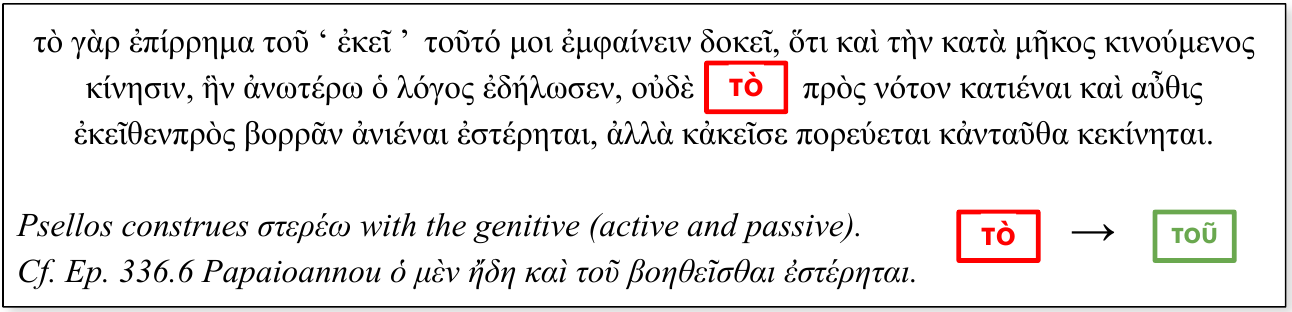}
    \caption{\textbf{Abridged dataset example}. The word \textgreek{τὸ} is labeled as an error (in this case scribal). The expert notes that Psellos uses the genitive with \textgreek{στερέω}, suggesting the text should read \textgreek{τοῦ}, and cites a parallel example from Papaioannou's edition of Letter 336.6 where Psellos uses \textgreek{τοῦ} with the same verb form. \autoref{sec:dataexample} provides the complete version of this example, and \autoref{sec:manuscript} includes an image of the manuscript showing how this scribal error may have been introduced.
}
    \label{fig:scribal_error}
\end{figure*}

\subsection{Labeling Process}
\label{sec:labeling}
The domain expert decides that a word is an error and gives the label $y=1$ for any of three reasons:
\begin{enumerate}
    \item \textit{Digitization Error (42 instances):} The expert confirms that the word in question is an error by comparing it with the corresponding text in the printed edition.
    \item \textit{Print Edition Error (114 instances):} The expert confirms that the word in question is an error by comparing it with the corresponding text in the available manuscripts.
    \item \textit{Scribal Error (61 instances):} The expert assesses the word in question to be a scribal error by philological reasoning.
\end{enumerate}
\autoref{fig:scribal_error} presents an abridged example from the dataset that contains a scribal error. For the manuscript referenced by the expert in identifying this scribal error, see \autoref{sec:manuscript}.
 We note that digitization and print errors can be identified with far greater confidence than scribal errors: for the latter, the assessments in the dataset must be considered preliminary only.

Not all words presented to the domain expert could be definitively labeled as real errors or not. In cases of potential scribal errors, where there is no explicit ground truth to verify an error and only reasoning based on textual evidence, the expert identified some words as possible errors, but not with sufficient confidence to label as $y=1$; a total of 237 such instances were labeled as either ``plausible'' or ``uncertain.''
We include these examples in the dataset but do not use them for evaluation purposes. Of the 763 words that were definitively labeled by the domain expert, 28\% were assigned the positive label $y=1$ (i.e., an error is present), while 72\% were assigned the negative label $y=0$ (i.e., no error is present).

\subsection{Impact of Over-Sampling}
The result of our sampling method is that all words presented to the domain expert, regardless of the label they receive, have a high CCR score (see \autoref{ccr}). To mitigate the distribution shift for non-errors ($y=0$) caused by over-sampling, we include a set of 237 randomly selected words from the corpus, assume they are non-errors due to the rarity of real errors, and assign them the label $y=0$. 

We note that this approach of over-sampling true positives is similar to that employed in computational methods for drug discovery, in which datasets are usually skewed toward drugs already likely to be effective, due to the similarly high cost of evaluation \citep{drugbank, compdrug, zagidullin2019drugcomb}. The case of computational drug discovery is similar in the sense that its goal is discovery---rather than scientific classification---and its bottleneck is in real-world evaluation, rather than computation. 

\subsection{Summary of the Dataset}

In summary, we used \citeposs{cowen-breen-etal-2023-logion} CCR metric to score a subset of words from the corpus of Michael Psellos, selecting the top 1,000 for expert review. The labeling process took over 100 hours and resulted in 763 words being definitively labeled. The remaining 237 words were labeled ``plausible'' or ``uncertain.'' 

The resulting dataset poses a challenging classification task, as many labels were determined through careful adjudication, consultation of source documents, and analysis of textual parallels. The classification task is made more challenging by the fact that the error detectors we consider have access to none of these materials.

\section{Deriving Error Detectors from LMs}\label{sec:scoring}

In this section, we describe the CCR metric and introduce two new error detection scoring metrics derived from LMs: $(1)$ the Pseudo Log-Likelihood Ratio (PLLR), originally developed for classification tasks in protein engineering, and $(2)$ discriminator scoring, using an ELECTRA discriminator without any additional fine-tuning. We also describe our methodology for prompting instruction-tuned LLMs to judge whether words are errors.\footnote{Future work should explore fine-tuning open-source LLMs on the task of error detection or posing it as a reward modeling task.}



\begin{table*}
\resizebox{\linewidth}{!}{
\begin{tabular}{llll}

\textbf{Model Type} & \textbf{Training Objective} & \textbf{Tokenization} & \textbf{Model Instance(s)} \\
\hline
\multirow{3}{*}{Encoder} & \multirow{3}{*}{Masked Language Modeling} & Character & BERT \\
 & & Sub-word & BERT (15\% \& 40\% mask ratio)\\
 & & Both & BERT \\
\cline{2-4}
 & \multirow{1}{*}{Replaced Token Detection} & Sub-word & ELECTRA\\
\hline
\multirow{2}{*}{Encoder-decoder} & \multirow{2}{*}{Span Corruption Denoising} & Character & T5\\
 & & Sub-word & T5\\
\hline
\end{tabular}
}
\caption{Suite of pre-trained models evaluated on error detection.}
\label{fig:models_table}
\end{table*}

\subsection{Chance-Confidence Ratio}
\label{ccr}
CCR is an error detector proposed by \citet{cowen-breen-etal-2023-logion} for the purpose of error detection and emendation. Given any MLM with learned conditionals $p(\cdot|\cdot)$, CCR scoring is defined as follows:
\[
T_{CCR}(\mathbf{w},i) = \frac{\max_{w\in\mathcal{W}^k_{w_i}} p(w|w_{-i})}{p(w_i|w_{-i})}
\]
where $\mathcal{W}^k_{w_i}$ denotes the set of words within Levenshtein distance $k$ of $w_i$, and $w_{-i}$ denotes the contextual sequence $\mathbf{w}$ with the entry at index $i$ masked. Intuitively, CCR is large when the \textbf{chance} of a word occurring in its given context, $p(w_i|w_{-i})$, is small relative to the \textbf{confidence} of the top model suggestion when restricted to Levenshtein distance $k$, $\max_{w\in\mathcal{W}^k_{w_i}} p(w|w_{-i})$. For dataset creation and all error detection experiments, we use $k=1$.

\subsection{Pseudo Log-Likelihood Ratio}
PLLR is a heuristic used by \citet{brandes2023genome} to predict whether a mutated protein sequence is malignant or benign. They find it to be an excellent zero-shot indicator of malignancy. PLLR takes a sequence and a mutated variant of that sequence and computes the ratio of the pseudo log-likelihoods of the sequence and its variant.

We propose applying PLLR to error detection by considering the hypothesis that each sequence of words in our text is itself a mutated variant of some original reference sequence, computing the score as follows:
\[
T_{PLLR}(\mathbf{w},i) = \frac{\max_{w\in\mathcal{W}^k_{w_i}} {\hat{p}(w_1,\hdots,w,\hdots,w_n)}}{{\hat{p}(w_1,\hdots,w_i,\hdots,w_n)}}
\]

Following \citet{brandes2023genome}, we compute pseudo-likelihood $\hat{p}(\cdot)$ with a single forward pass of a MLM by multiplying the probabilities of the ground-truth token at each output position, taking advantage of the fact that MLMs output a probability distribution at all positions. While this approach is highly heuristic, computing $\hat{p}(\cdot)$ is efficient insofar as it requires only a single forward pass.

\subsection{Discriminator Scoring}
 We additionally propose using a discriminator model for binary classification on each token to predict whether it is the original or a replacement sampled from a generator. This aligns closely with the phenomenon that we are attempting to model, where a scribe, acting as a generator, occasionally alters words in a text.

\subsection{Few-Shot Prompting}
Although today's instruction-tuned LLMs are not specifically designed for tasks involving premodern Greek, their training on extensive internet crawls suggests that they could encounter some relevant data \citep{achiam2023gpt, touvron2023llama}. We provide sequences of premodern Greek and ask the instruction-tuned LLM to assess whether a specified word is an error, giving examples with expert annotations. We prompt the LLM to return a score from $1$ to $m$ indicating how likely a given word is to be an error.\footnote{We try $m=2, 3, 5, 10$ and find $m=5$ to be best.} More prompting details are made available in our source code.

\section{Overview of LM Pre-Trainings}
Each error detector we evaluate is unsupervised, using distributions from language model pre-training objectives rather than being trained on a labeled error dataset. Crucially, we pre-train all models from scratch, avoiding existing premodern Greek models to prevent contamination between their training data and our dataset.\footnote{Note, however, that we have no such assurances about the training data used for GPT-3.5 and GPT-4.} Our goal is to compare error detection methods, not specific models, which vary in data, compute, and parameters. To ensure a fair comparison, we keep these factors as consistent as possible across the seven models we pre-train.

\begin{figure*}
    \centering
    \includegraphics[width=1.0\linewidth]{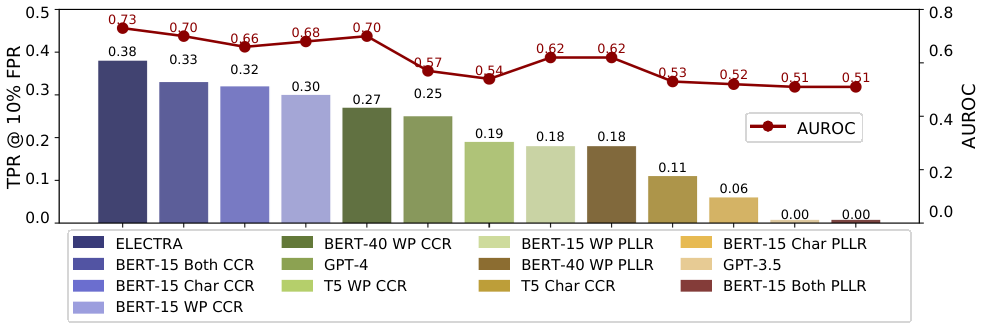}
    \caption{\textbf{AUROC and TPR at 10\% FPR for each error detector}. ``15'' and ``40'' refer to mask ratios, ``Char'' and ``WP'' refer to character and sub-word tokenization, and ``Both'' refers to the combined tokenization method.}

    \label{fig:auroc_tpr}
\end{figure*}

 \begin{table}[]
    \centering
    \begin{tabular}{|l||c|c|c|}
    \hline
         \multirow{2}{*}{Model}& \multicolumn{3}{c|}{Type of error} \\
        \cline{2-4}
         & Digitization & Print & Scribal \\
        \hline
        ELECTRA & 0.75 & 0.71 & \textbf{0.59} \\
        BERT (Best) & 0.65 & 0.67 & \textbf{0.57} \\
        T5 (Best) & 0.61 & 0.53 & \textbf{0.52} \\
        GPT-4 (Best) & 0.53 & \textbf{0.52} & \textbf{0.52} \\
        \hline
    \end{tabular}
\caption{AUROC of select detectors when $y=1$ examples are limited to specific error categories. Scribal errors are universally the most challenging (in bold). ``Best'' refers to the highest-AUROC detector of each model type.}
    \label{tab:error_types}
\end{table}

\subsection{Pre-Training Data}
We assemble pre-training data from sources made available by prior work, including \citet{singh-etal-2021-pilot}, \citet{cowen-breen-etal-2023-logion}, and \citet{riemenschneider-frank-2023-exploring}. We divide the training, validation, and testing splits so that no exact 50-character overlap in training occurs in validation or testing. In total, our training set contains about 120M words of premodern Greek. We do not remove redundancies within the training split. We do, however, exclude all texts in the corpus of Michael Psellos, ensuring that the dataset remains fully held-out from all model trainings. 

\subsection{Tokenization}
\label{sec:tokenization}
Since error detection requires sensitivity to character-level changes in text, it is possible that prevalent sub-word tokenization methods such as Byte-Pair Encoding \citep{sennrich2015neural} and WordPiece \citep{schuster2012japanese} are sub-optimal for the task. To investigate this, we pre-train models with both a WordPiece tokenizer with a vocabulary size of 50K and a character-level tokenizer. Following \citet{assael2022restoring}, we additionally train a character-level BERT model with an auxiliary sub-word embedding table, with the aim of incorporating different token granularities for prediction. Although different models utilize different tokenizers, we standardize training examples to contain identical text for each. Specifically, we maximally stack consecutive sentences until the number of character-level tokens exceeds 1,024.

\subsection{Pre-Training Configurations}

We train several variations of bidirectional encoder or encoder-decoder models as listed in \autoref{fig:models_table}.
 These include four BERT models: three models with 15\% and 40\% mask ratios using a sub-word tokenizer, and a 15\% mask ratio using a character-level tokenizer.\footnote{\citet{wettig-etal-2023-mask} suggest that a 40\% mask ratio is superior to 15\% for uniform masking.} The fourth is a custom character-level BERT integrated with an auxiliary sub-word embedding table. Additionally, we pre-train two T5 models \citep{raffel2020exploring}, one each with sub-word and character-level tokenizers. Finally, we pre-train an ELECTRA discriminator in tandem with a generator which we later discard. We train each model on four A100 GPUs for six days or until validation loss converges. For full model training parameters, see \autoref{sec:hyperparameters}.
 

\section{Evaluation}

An error detector $T$ is evaluated by the quality of its predictions $T(\mathbf{w},i)=\hat{y}$ on labeled data. For evaluation purposes, we treat $T$ as a binary classifier which declares $w_i$ to be an error when $T(\mathbf{w},i)\geq t$ for a fixed threshold $t\in\mathbb{R}$. We compare error detectors based on their true positive rate (TPR) at a fixed false positive rate (FPR), as seen in \autoref{fig:auroc_tpr}. We also consider AUROC, defined to be the area under the graph consisting of pairs of FPRs and TPRs over all $t\in\mathbb{R}$.

\subsection{Computing Error Scores}
 We use BERT and T5 models for computing CCR scores, BERT models for PLLR scores, ELECTRA for discriminator scores, and GPT-3.5 and GPT-4 for few-shot prompting scores.\footnote{We use \texttt{gpt-3.5-turbo} and \texttt{gpt-4-1106-preview} with a temperature of $1.0$.} We evaluate these error detectors on 763 labeled examples from our dataset and 237 randomly sampled words from the corpus that are presumed to be non-errors. 
 \begin{figure}
    \centering
    \includegraphics[width=0.5\textwidth]{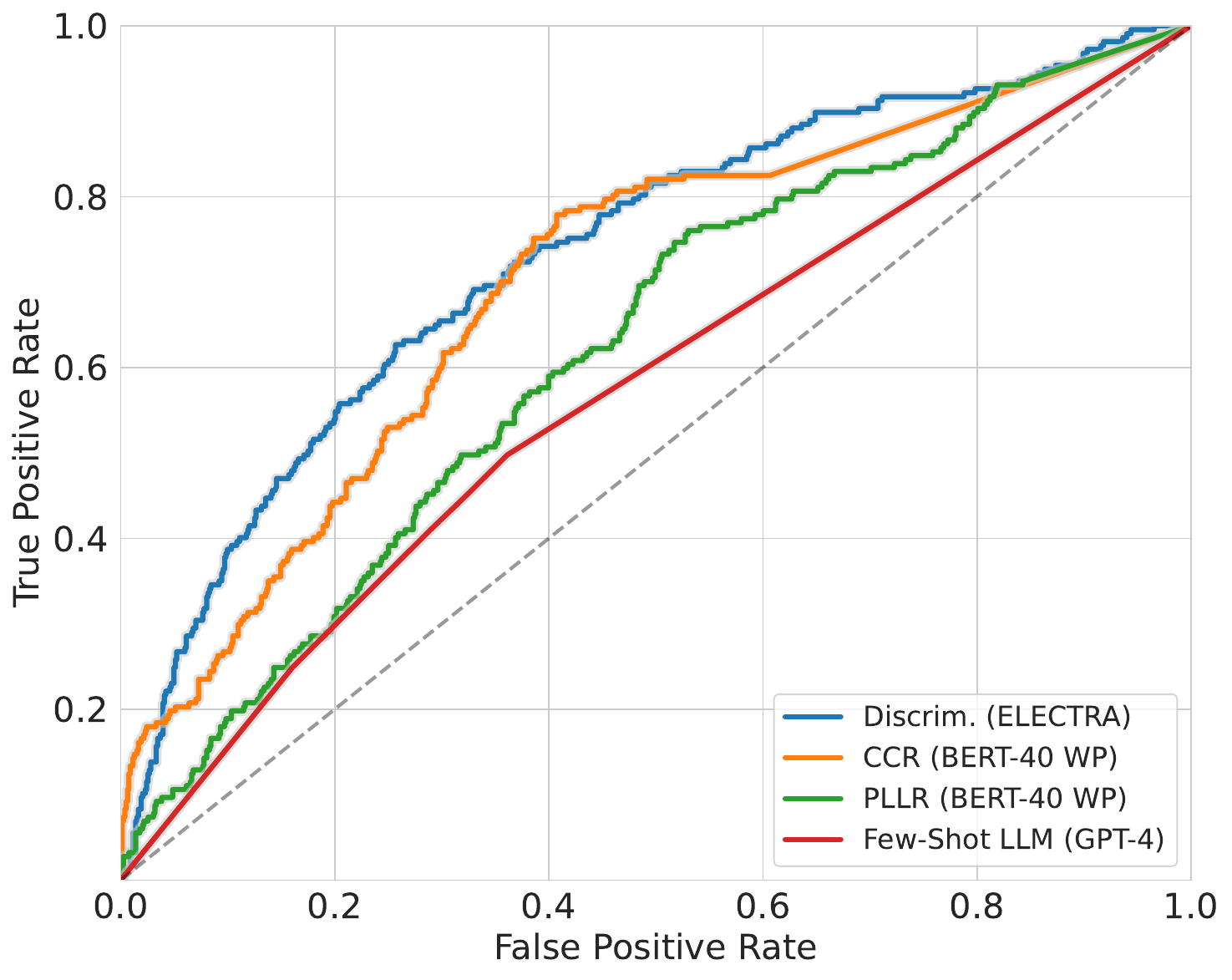}
    \caption{ROC curves of the best performing error detectors of each type. BERT-40 WP denotes the sub-word BERT model trained with 40\% mask ratio.}
    \label{fig:alldata}
\end{figure}
 
 \subsection{Results}
 The ELECTRA-based error detector achieves the highest scores in both TPR at 10\% FPR and AUROC, marking a new state-of-the-art on the classification task introduced with our new dataset. The four BERT-based CCR error detectors are the next best performing in both metrics. In comparison, PLLR-based detectors, T5-based CCR detectors, and few-shot prompted LLMs are noticeably less effective. 
 
 Considering the best-performing detector from each category, we observe a clear ranking, as illustrated by the ROC curves in \autoref{fig:alldata}: Discriminator Scoring is best, followed by CCR, then PLLR, then Few-Shot LLM Prompting. The results do not provide a strong signal for which tokenization method is best. Extended comparisons across models and methods can be found in \autoref{sec:additional_curves}. 

\begin{figure}
    \centering
    \includegraphics[width=0.49\textwidth]{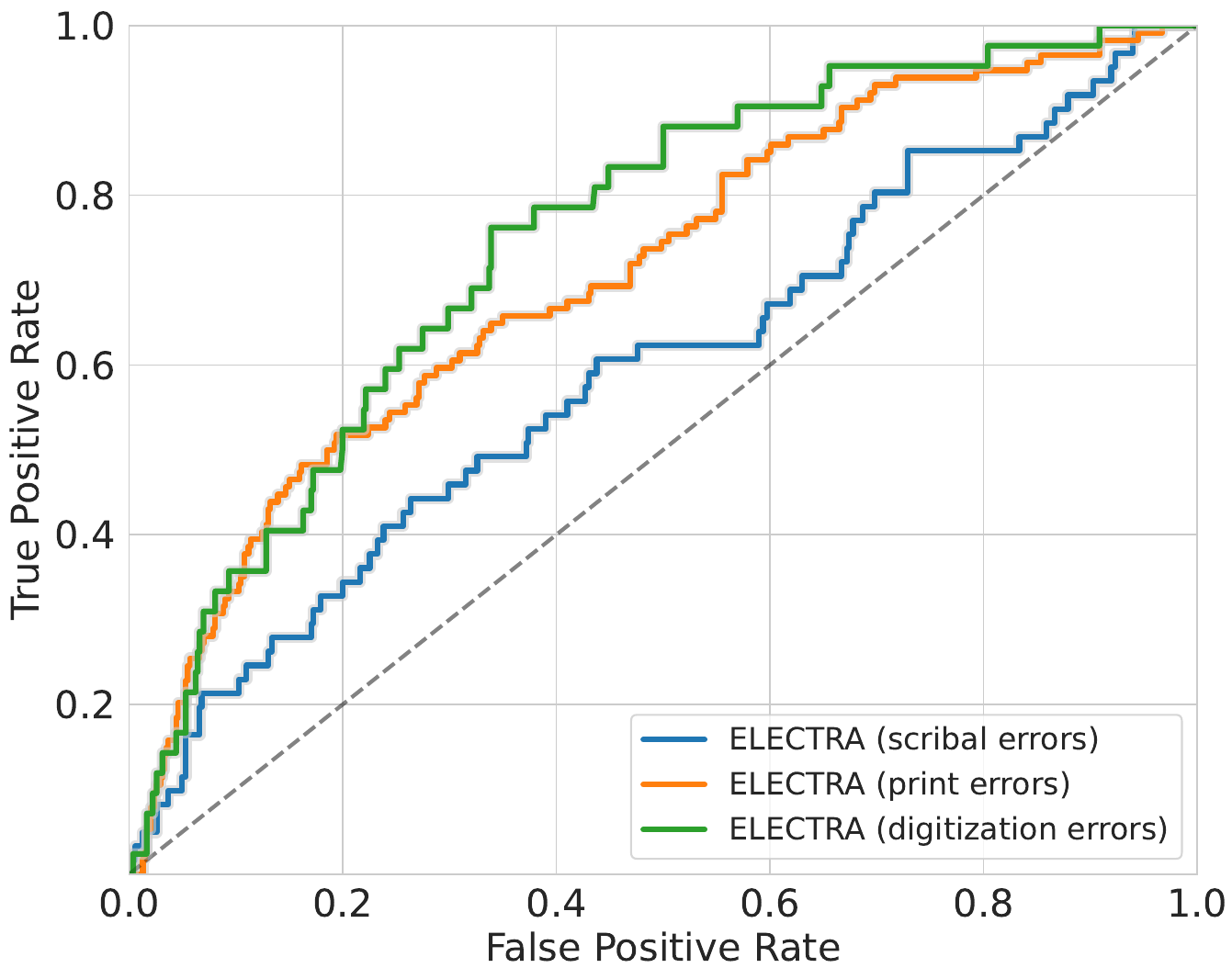}
    \caption{ROC curves of ELECTRA across types of errors.}
    \label{fig:electra_categorized}
\end{figure}

 Moreover, we observe across methods that scribal errors are more challenging to detect than print and digitization errors. \autoref{tab:error_types} shows that the best-performing detectors of each model type have the lowest AUROC scores for classifying scribal errors. For ELECTRA and BERT-based CCR, which are the most effective error detectors, the drop is especially pronounced. \autoref{fig:electra_categorized} shows this phenomenon for ELECTRA, with ROC curves corresponding to different error types clearly separated. AUROC scores on scribal errors for all models hover relatively close to the random baseline of $0.5$. The ease of detecting errors correlates with the recency of the stage in which they were introduced.

\section{Discussion}
The superior performance of ELECTRA as an error detector on our newly created dataset has important implications for machine learning-assisted error discovery. Until now, unsupervised error detection in premodern texts has only employed BERT-based CCR. However, our results indicate that discriminator-based models, like ELECTRA, outperform CCR when evaluated on real copying errors. That said, there are still advantages to using BERT-based models: for a given index, $\arg\max_{w\in\mathcal{W}^k_{w_i}} p(w|w_{-i})$ produces a suggested token within a specified Levenshtein distance, enabling error correction in addition to detection. Future work in error correction could leverage a generator alongside the discriminator to a similar effect.

 ELECTRA's success is, in some ways, surprising: the method of over-sampling words with high CCR scores to create this dataset creates a bias for words with a \textit{low} chance metric (see \autoref{ccr}); on the other hand, the ELECTRA discriminator is primarily trained to detect erroneous tokens with \emph{high} chance values, as they are sampled directly from a generator.\footnote{ELECTRA learns to sometimes propose the \textit{lectio difficilior}, whereas error detectors guided by chance propose the \textit{lectio facilior}, to employ the terminology of philological scholarship.} Among other considerations, future work could restrict the generator to sample only from $\mathcal{W}_{w_i}^k$ to better simulate the distribution of real errors.

Despite a marked improvement in TPR from GPT-3.5 to GPT-4, both models struggle to classify words effectively, with AUROC scores of $0.51$ and $0.57$, respectively. Both models produce seemingly well-reasoned yet ultimately misinformed explanations for their classifications. In one telling reply, GPT-3.5 rationalizes a $5/5$ error score as follows:

\begin{quote}
``The word `\textgreek{σαφες}' is indeed an error. The correct form should be `\textgreek{σαφης},' as it should agree with the neuter noun `\textgreek{το πραγμα}' in the nominative singular form. The ending \textgreek{-ες} is masculine, while \textgreek{-ης} is the proper form for a neuter adjective in this context. This is a clear grammatical error that needs correction.''
\end{quote}
The word in question is, in fact, correct and GPT-3.5's explanation disregards basic rules of Greek grammar. We cannot blame this particular lapse on the contamination of modern data, as \textgreek{σαφές} remains a neuter form in Modern Greek.

We also note the relative under-performance of the proposed PLLR metric. During experiments, we observe that the words maximizing a sequence's pseudo-likelihood are often nonsensical. It appears that adding noise in one position of a sequence can counterintuitively bolster the ground-truth logits occurring in other positions in this pseudo-likelihood setting.

\section{Conclusion}
We present the first annotated dataset of real errors in premodern Greek texts with a view to improving the evaluation of error detection. We propose new error detection methods and evaluate them on the new dataset using an array of pre-trained models, including different configurations of BERT and T5, ELECTRA, and instruction-tuned LLMs like GPT-4. We find that our proposed discriminator-based detector outperforms other methods and establishes a state-of-the-art for the error detection task introduced by our new dataset. Additionally, we observe across methods that scribal errors are more challenging to detect than print and digitization errors.

Our dataset serves as an important new resource for evaluating the efficacy of machine learning methods in detecting real errors in premodern texts and offers a benchmark for the development of more effective error detection algorithms. Evaluating error detection methods on real errors paves the way for accelerated error discovery and machine-learning assisted restoration of premodern texts. We hope that by creating this dataset and presenting new error detection methods, we can introduce an iterative cycle of improvement, where better datasets lead to better detectors, which in turn lead to even better datasets, and so forth.

\section*{Limitations}
Models like BERT, ELECTRA, and T5 are traditionally pre-trained and then fine-tuned for specific tasks. In our case, we employ these models directly from pre-training for error detection, which leads to misalignment with their original training objectives. For instance, while the standard MLM task masks about 15\% of tokens (roughly 75 tokens in a 500-token example), error detection methods like CCR and PLLR can involve masking just one token at a time, thus resulting in an input that is out of distribution.\footnote{A training adjustment to alleviate this effect could be a decaying mask-ratio scheduler.} In this study, we aim to better understand the use of pre-trained language models in the zero-shot setting of error detection scoring. 

The circularity of dataset creation and error-detector evaluations is a legitimate concern. Due to the very slow pace (up to many hours per datapoint) of annotation, there is no other known option than to oversample likely errors in some way. Moreover, we note that although the labeled words are oversampled using the BERT CCR metric, the ELECTRA-based detector outperforms the BERT CCR detectors. It is our hope that this dataset will spark the development of better error detectors than those we present here, and that those will yield datasets of their own, which may be cross-referenced against ours to measure the legitimacy of this concern.

Furthermore, our dataset is limited to 1,000 words from a single author. It is restricted in both size and scope due to the significant demands that generating it places on domain experts. We focus on the end task of error detection and deliberately omit examining the relationship between different manuscript copies.

\section*{Ethics Statement}
 Pre-training language models is computationally intensive. As we focus on an underrepresented language, we hope that the models and methods we produce will serve as valuable resources for the scholarly community, with utility extending beyond the scope of this paper.

 \section*{Acknowledgements}
 We thank Anirudh Ajith, Julia Balla, Jack Geld, Mirjam Kotwick, Anika Maskara, and Howard Yen for their valuable feedback and advice. We gratefully acknowledge funding from a Magic Grant awarded by the Princeton Humanities Council and computational resources provided by Princeton Language and Intelligence (PLI).

\bibliography{anthology,custom}
\clearpage
\onecolumn
\begin{center}
\Large \textbf{Appendix}
\end{center}
\appendix
\vspace{0.5cm}
\section{\Large Manuscript Section Containing Scribal Error}
\label{sec:manuscript}

\begin{figure*}[ht]
    \centering
    \includegraphics[width=\textwidth]{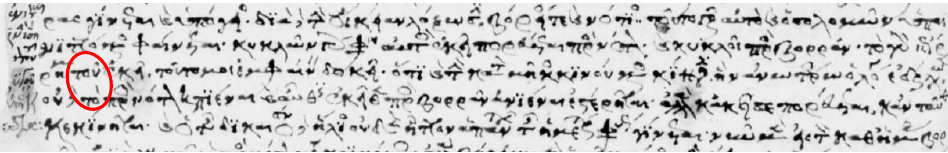}
    \caption{Manuscript section (Cod. Paris. gr. 1182, f. 26v) containing text discussed in \autoref{sec:dataexample}.}
    \label{fig:wide_image}
\end{figure*}

\noindent In the figure above, within the red oval, we see \textgreek{τοῦ} (on top) and \textgreek{τὸ} (below), corresponding to \textgreek{τοῦ} in \textgreek{τοῦ ἐκεῖ} and \textgreek{τὸ} in \textgreek{τὸ πρὸς νότον} from the snippet of text in \autoref{fig:scribal_error} and \autoref{sec:dataexample}. The BERT-based CCR detector flagged \textgreek{τὸ} as an error, which the domain expert determined to be a scribal error based on textual parallels and Psellos’s usage of the verb \textgreek{στερέω}. Upon further review of the manuscript, the expert noted that this error is connected to another mistake in the line just above: \textgreek{τοῦ} in the line above (also within the red oval) should read \textgreek{τὸ}. The proximity and similarity of these two words likely caused the confusion. 

\vspace{1cm}

\section{\Large Model Training Hyper-Parameters}\label{sec:hyperparameters}
While the remaining weights are initialized randomly, we initialize the embedding table of the ELECTRA discriminator using a pre-trained BERT model. 
We train the ELECTRA generator from scratch in tandem with the discriminator. Preliminary testing showed that using a pre-trained generator, even with a temperature schedule (cf. \citet{dong2023fast}), hindered the discriminator's learning. 
For the character-level BERT model with an auxiliary sub-word embedding table, we use \citeposs{devaul_desformers} implementation, which is a fork of HuggingFace's \texttt{BertForMaskedLM} module. 
\begin{table}[h]

     \centering

     \begin{tabular}{llll}
         \toprule
         \textbf{Hyperparameter} & \textbf{BERT} & \textbf{ELECTRA} & \textbf{T5}  \\
         \midrule
         Attention Heads &12 &12& 12\\
         Per Device Batch Size &16&16&16\\
         Hidden Dropout &0.1&0.1&0.1\\
         Hidden Size &768&768&768\\
         Learning Rate (LR) &$5\cdot 10^{-5}$&$5\cdot10^{-5}$& $1\cdot 10^{-4}$\\
         LR Scheduler &linear& linear &cosine\\
         Nb.\ of Layers &12 &12& 2 $\cdot$ 12\\
         Warmup Steps &0&0&10000\\
         \bottomrule
     \end{tabular}
      \caption{Hyper-parameter settings for model training. Our experiments involve two types of models: those utilizing a 50,000-token subword vocabulary and those with character-level input. The remaining hyper-parameters are unchanged from the corresponding HuggingFace model configurations.} 
     \label{tab:my_label}
 \end{table}

\clearpage

\section{\Large CCR Implementation Details}
\label{sec:implementation}
\vspace{0.25cm}
As detailed in \autoref{sec:scoring}, our evaluation requires that each experiment produces a list of scores $t \in \mathbb{R}$, corresponding directly to the list of ground truth labels $y \in \{0,1\}$.

\noindent As words can consist of multiple sub-word tokens, in practice we calculate CCR (\autoref{ccr}) for $w_i$ with tokens $t_1, ..., t_n$ with the following heuristics for chance and confidence: 

\[
\text{chance} \gets \min_{j=1}^n p(t_j|t_{-j})
\]
For confidence, we replace masking $w_i$ with 1 to $n$ mask tokens and beam search across each masked sequence to find the top suggestions within Levenshtein distance $k$ of $w_i$. The confidence is determined as:
\[
\text{confidence} \gets \max_{m=1}^n \left(\max_{w'\in\mathcal{W}^k_{w_i}} p(w'|w_{-m})\right)\,,
\]
where $w_{-m}$ indicates the sequence with $w_i$ replaced by $m$ mask tokens. We use a beam size of 10, and if beam search cannot find any $w'$ within distance $k$ of $w_i$, we return a score of 0. With BERT and T5 models, we compute \( p(\cdot | t_{-i}) \) by inserting a masked token at position \( i \) and then applying softmax to the logits at position \( i \).
\\ \\
Computing $p(\cdot | w_{-i})$ with BERT is straightforward: simply replace the token at position $i$ with a mask token and perform a forward pass to obtain the desired distribution. With T5, this computation is more heuristic: instead of directly replacing a single token, a span corruption approach is used where a token at position $i$ is replaced with the placeholder <extra\_id\_0>. We then make use of the distribution of potential spans produced by a forward pass.

\vspace{1cm}

\section{\Large Dataset Example}\label{sec:dataexample}
\vspace{0.25cm}
\begin{description}
    \item[\large Transmitted Word in Question:] \textgreek{τὸ}
    \item[\large Expert Label:] GOOD FLAG.
    \item[\large Model-Suggested Alternative:]     \textgreek{του}
    \item[\large Further Expert Notes:] \hphantom{d}\\[0.5em]
    GOOD FLAG. GOOD SUGGESTION. Scribal. Codex unicus. Corrupt.\\[0.5em]
    MS P. Psellos construes στερέω with the genitive (active and passive). The error appears to be related to a further corruption earlier in the same sentence, which the error detector did not identify: for transmitted \textgreek{τοῦ} ` \textgreek{ἐκεῖ} ' read \textgreek{τὸ} ` \textgreek{ἐκεῖ} ' and note the position of \textgreek{τοῦ} < \textgreek{τὸ} immediately above \textgreek{τὸ} < \textgreek{τοῦ} in the relevant manuscript (Cod. Paris. gr. 1182, f. 26v).\\[0.5em]
    1. Michael PSELLUS Epist., Hagiogr., Phil., Polyhist. et Theol. Theologica {2702.012} Opusculum 107 line 56\\
    \textgreek{ρημα τοῦ ‘ἐκεῖ’ τοῦτό μοι ἐμφαίνειν δοκεῖ, ὅτι καὶ τὴν κατὰ μῆκος κινού- (55)\\
    μενος κίνησιν, ἣν ἀνωτέρω ὁ λόγος ἐδήλωσεν, οὐδὲ τὸ πρὸς νότον κατιέναι\\
    καὶ αὖθις ἐκεῖθεν πρὸς βορρᾶν ἀνιέναι ἐστέρηται, ἀλλὰ κἀκεῖσε πορεύεται}\\
    \item[\large Word Index in Text:] 27
    \item[\large Text:] \hphantom{d}\\[0.5em]
    \textgreek{τὸ γὰρ ἐπίρρημα τοῦ ’ ἐκεῖ ‘ τοῦτό μοι ἐμφαίνειν δοκεῖ , ὅτι καὶ τὴν κατὰ μῆκος κινούμενος κίνησιν , ἣν ἀνωτέρω ὁ λόγος ἐδήλωσεν , οὐδὲ τὸ πρὸς νότον κατιέναι καὶ αὖθις ἐκεῖθεν πρὸς βορρᾶν ἀνιέναι ἐστέρηται , ἀλλὰ κἀκεῖσε πορεύεται κἀνταῦθα κεκίνηται . Καὶ ’ ὁ τῆς δικαιοσύνης ‘ δὲ ’ ἥλιος ‘ οὐδὲν ἧττον ἁπανταχοῦ τῆς ἡμετέρας φύσεως γίνεται , νῦν μὲν εἰς τὸν καθ´ ἡμᾶς βορρᾶν ἀνιών , νῦν δὲ πρὸς νότον μετακλινόμενος . ἀλλὰ βόρειον μὲν ἡμῖν μέρος πρὸς ὕψος ἠρμένον καὶ πολλαῖς μοίραις τῆς γῆς μετεωριζόμενον ὁ κοσμῶν νοῦς τὴν ψυχήν · νότιον δὲ ἡ μετέχουσα τοῦ νοῦ ψυχή , ὑποβεβηκυῖα μὲν ἐκεῖνον καὶ κάτω ποι τεταγμένη , οὐδ´ αὐτὴ δὲ ἀμοιροῦσα τοῦ θείου φωτός . ἢ βορρᾶς μὲν ἡμῖν τὸ σύμπαν νοητόν , ὅσον τε ἐν νῷ καὶ ὅσον ἐν τῇ ψυχῇ , νότος δὲ τὸ συμπεριειλημμένον τῇ ὕλῃ σῶμα , μᾶλλον δὲ τὸ ταύτην συμπεριλαβόν . ἔμελλε γὰρ ἡ καθ´ ἡμᾶς ὕλη ὅσον ἐπὶ τῇ οἰκείᾳ φύσει ἀμέτοχος εἶναι καλοῦ , ἀλλ´ ὁ πορευόμενος πρὸς νότον καὶ κυκλῶν πρὸς βορρᾶν οὐδὲ ταύτην ἀποστερεῖ τῶν οἰκείων μαρμαρυγῶν , οὐ μόνον οἷς ἐπιτηδείαν ἐργάζεται πρὸς εἴδους καταδοχήν , οὐδ´ ὅτι ὁμοῦ τε ὑπέστησε καὶ πρὸς τὴν κοσμοποιίαν ἐχρήσατο , ἀλλ´ ὅτι καὶ τὰ πολλὰ τῶν πρακτικῶν ἀρετῶν διὰ ταύτης κατορθοῦσθαι εἴωθεν , εἴπερ αἱ μὲν δέονται σώματος , τὸ δὲ ὕλης οὐκ ἄτερ . Εἶτα πῶς οὐκ ἐσκότωνται οἱ μὴ τὸν τοῦ πατρὸς λόγον κυρίως θεὸν ὀνομάζοντες , δι´ οὗ καὶ τὸ θεοῦσθαι τοῖς θεουμένοις ἐστίν , ἀλλὰ τὴν μὲν γέννησιν ἀπαρνούμενοι , ἵνα μὴ πάθος εἰσαγάγωσι , τὴν δὲ κτίσιν αὐτοὶ ἀναπλάττοντες , ἵν´ ὁμόδουλον ἡμῖν τὸν δημιουργὸν ποιήσωσιν ; εἰσὶ δὲ οἳ προσίενται μὲν τὴν γέννησιν , ὥσπερ δὴ καὶ τὴν ἀγεννησίαν , οὐσίας δὲ ταύτας ἀντιδιῃρημένας φασίν , ὥσπερ τὸ σῶμα καὶ τὸ ἀσώματον , καὶ θεὸν μὲν ἑκατέραν τῶν οὐσιῶν λέγουσιν , ἀκυρίαν δὲ καὶ ὁμωνυμίαν προσάπτουσι τοῖς μόνοις κυρίοις καὶ ὑπὲρ πᾶσαν λογικὴν μέθοδον . Πρὸς οὓς ὁ μέγας πατὴρ ἀπαντῶν ’ ὁ μὲν οὖν ἡμέτερος ‘ φησί ’ λόγος ὥσπερ ἵππου καὶ βοὸς καὶ ἀνθρώπου καὶ ἑκάστου τῶν ὑπὸ τὸ αὐτὸ εἶδος εἷς λόγος ἐστί · καὶ ὃ μὲν ἂν μετέχῃ τοῦ λόγου , τοῦτο καὶ κυρίως λέγεσθαι , ὃ δ´ ἂν μὴ μετέχῃ , τοῦτο μὴ λέγεσθαι ἢ μὴ κυρίως λέγεσθαι .}\\
\end{description}





\clearpage

\section{\Large Additional ROC Curves}\label{sec:additional_curves}

\begin{figure}[h!]
    \centering
    \begin{minipage}[b]{0.45\textwidth}
        \centering
        \includegraphics[width=\textwidth]{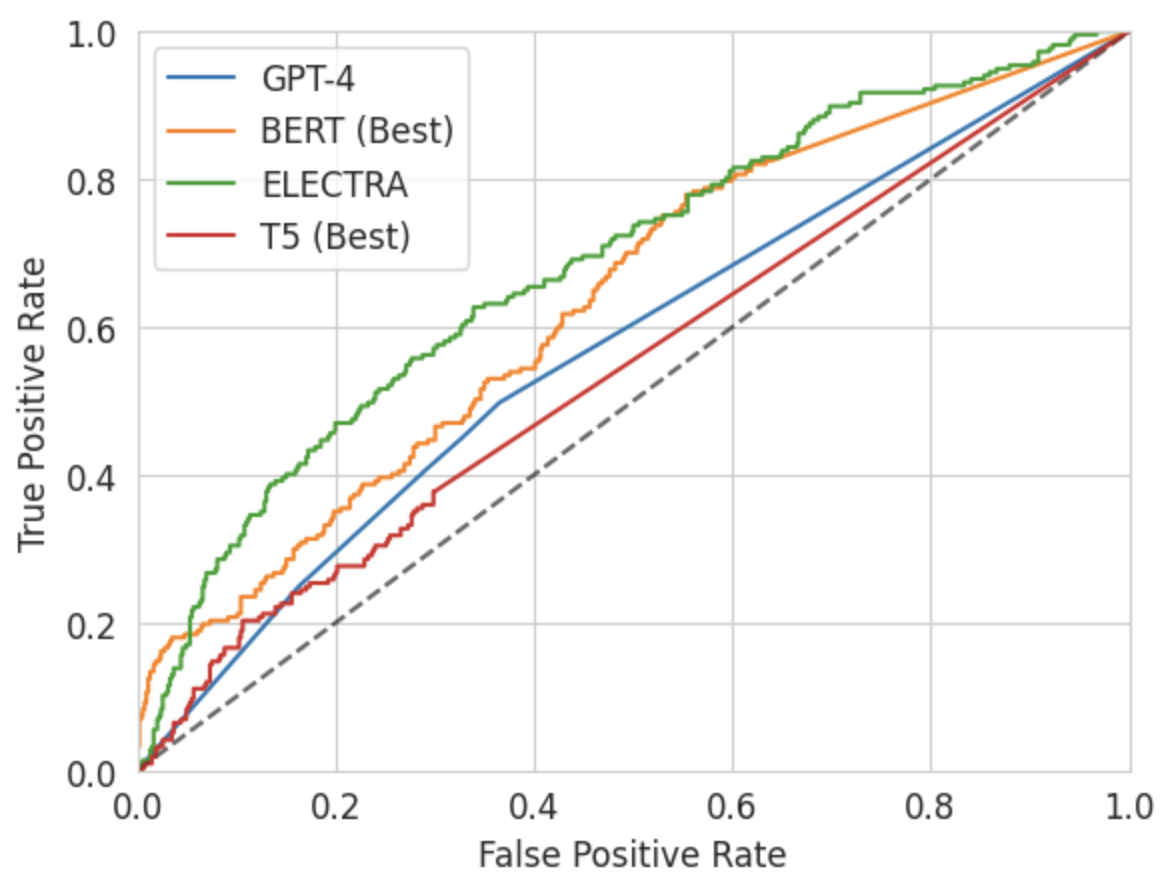}
        \caption{ROC curves of the best performing error detectors of each model type excluding the 237 presumed non-errors sampled from the corpus.}
        \label{fig:humandata}
    \end{minipage}
    \hfill
    \begin{minipage}[b]{0.45\textwidth}
        \centering
        \includegraphics[width=\textwidth]{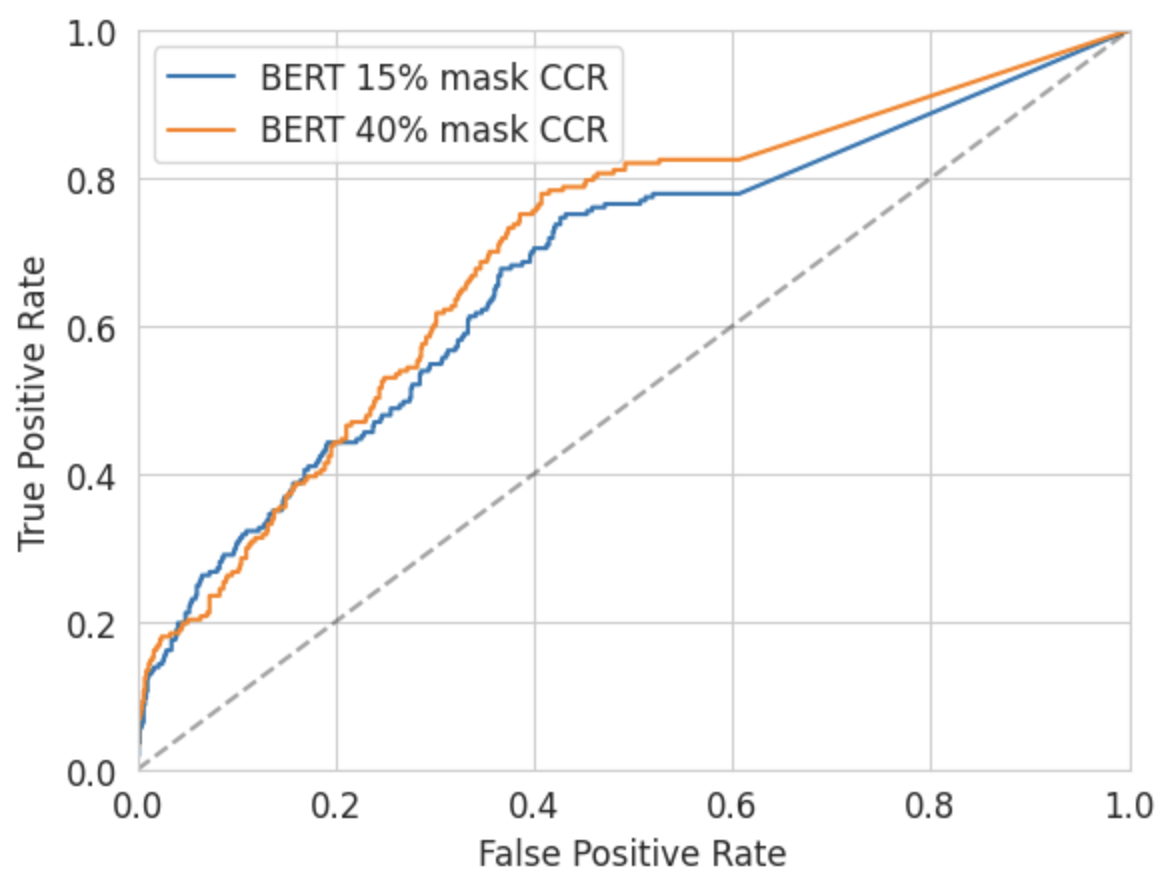}
        \caption{Comparison of ROC curves for BERT models trained with different mask ratios}
        \label{fig:bert_40_15}
    \end{minipage}
    
    \vspace{0.5cm} 
    
    \begin{minipage}[b]{0.45\textwidth}
        \centering
        \includegraphics[width=\textwidth]{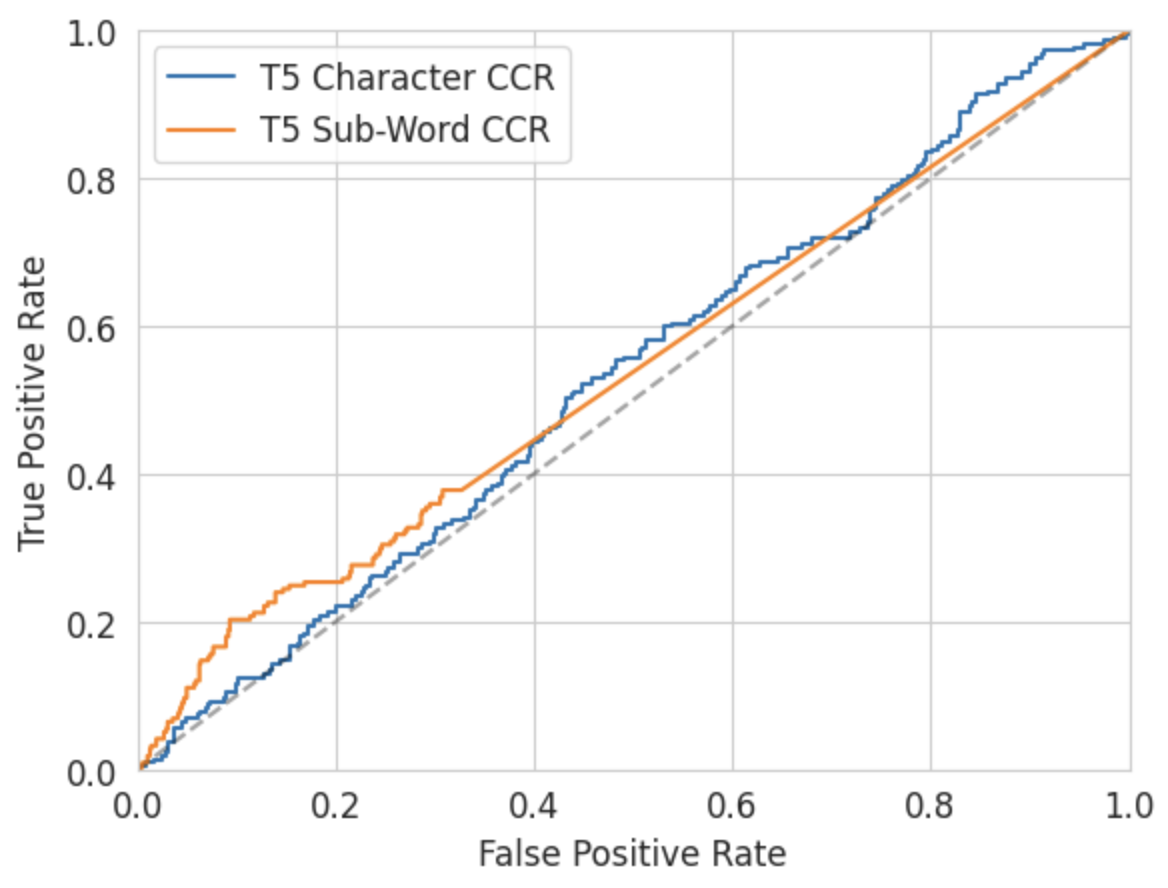}
        \caption{Comparison of ROC curves for T5 models trained with different tokenizers.}
        \label{fig:t5_comparison}
    \end{minipage}
    \hfill
    \begin{minipage}[b]{0.45\textwidth}
        \centering
        \includegraphics[width=\textwidth]{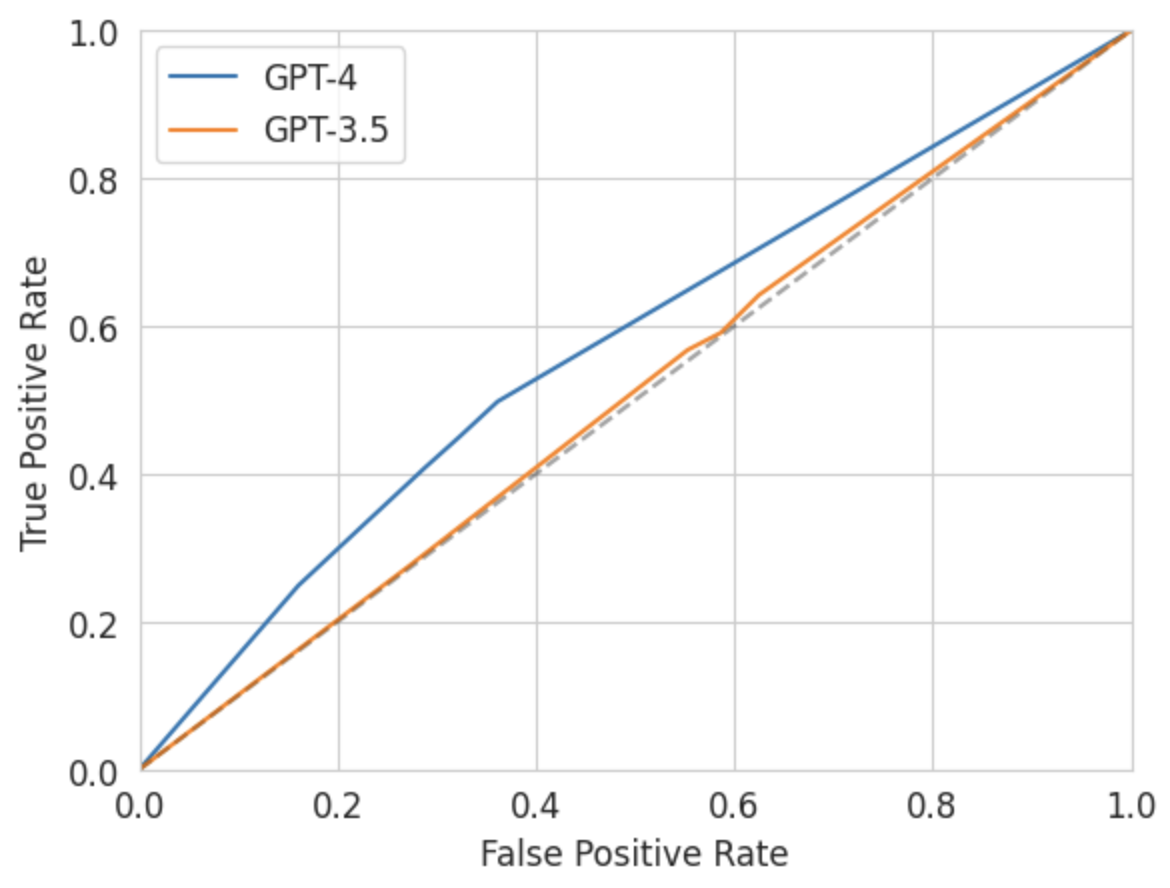}
        \caption{Comparison of ROC curves for GPT-3.5 and GPT-4.}
        \label{fig:gpts}
    \end{minipage}
\end{figure}

\end{document}